# Characteristics Analysis of Autonomous Vehicle Pre-crash Scenarios

Yixuan Li[a], Xuesong Wang[†a,b], Tianyi Wang[c], Qian Liu[a]

[a]*School of Transportation Engineering, Tongji University, No. 4800 Cao'an Road, Shanghai 201804, China*
[b]*The Key Laboratory of Road and Traffic Engineering, Ministry of Education, Shanghai 201804, China*
[c]*Department of Mechanical Engineering & Materials Science, Yale University, 17 Hillhouse Avenue, New Haven, CT 06511, United States*

**Abstract**

To date, hundreds of crashes have occurred in open road testing of automated vehicles (AVs), highlighting the need for improving AV reliability and safety. Pre-crash scenario typology classifies crashes based on vehicle dynamics and kinematics features. Building on this, characteristics analysis can identify similar features under comparable crashes, offering a more effective reflection of general crash patterns and providing more targeted recommendations for enhancing AV performance. However, current studies primarily concentrated on crashes among conventional human-driven vehicles, leaving a gap in research dedicated to in-depth AV crash analyses. In this paper, we analyzed the latest California AV collision reports and used the newly revised pre-crash scenario typology to identify pre-crash scenarios. We proposed a set of mapping rules for automatically extracting these AV pre-crash scenarios, successfully identifying 24 types with a 98.1% accuracy rate, and obtaining two key scenarios of AV crashes (i.e., rear-end scenarios and intersection scenarios) through detailed analysis. Based on the abundance of crash data, we adopted different analysis methods to analyze the features of key scenarios. Association analyses of rear-end scenarios showed that the significant environmental influencing factors were traffic control type, location type, light, etc. For intersection scenarios prone to severe crashes with detailed descriptions, we employed causal analyses to obtain the significant causal factors: habitual violations and expectations of certain behavior. Optimization recommendations were then formulated, addressing both governmental oversight and AV manufacturers' potential improvements. The findings of this paper could guide government authorities to develop related regulations, help manufacturers design AV test scenarios, and identify potential shortcomings in control algorithms specific to various real-world scenarios, thereby optimizing AV systems effectively.

*Keywords:* Automated vehicle crash, Pre-crash scenario typology, Safety assessment, Association analysis, Causal analysis

## 1. Introduction

Autonomous driving technology is becoming increasingly crucial in automotive and traffic engineering worldwide. Autonomous vehicles (AVs) utilize a perception-decision-execution framework, offering great potential to improve traffic efficiency and safety. However, despite rapid technological advancements, substantial challenges persist (Lee and Yoon (2024); Lin and Wang (2022)). Research has shown that while autonomous driving aims to achieve "Vision Zero", open road testing of AVs highlights great safety gaps. The average mileage between crashes for conventional vehicles is approximately 500,000 miles, whereas for AVs, it is only 42,017 miles (Favaro and Varadaraju (2017)). AV crashes involving serious injuries or fatalities, such as incidents with Tesla, Google, and Uber, have provoked widespread concern and debate. Consequently, improving the safety of AVs has become one of the key research priorities in this field.

AV open road testing, conducted on shared roads with other vehicles and road users, provides a more comprehensive and realistic evaluation of the performance of AVs in complex scenarios. Data on crashes, loss of control, detachment, and other AV-related incidents during testing contribute to the analysis of AV system shortcomings. Among these, crash data are particularly crucial, which serve as the foundation and prerequisite for identifying key factors influencing AV performance and developing optimization solutions to improve the reliability, stability, and safety of AVs.

The United States has taken the lead in advancing AV open road testing, beginning in 2014 when the California Department of Motor Vehicles(CA DMV) initiated its AV road testing program, which requires AV manufacturers to submit detailed AV crash reports (Administration (2024)). By January 2024, 618 crash reports involving Level 2 to Level 4 AVs had been published (of Motor Vehicles (CA DMV)). These comprehensive crash reports serve as a reliable resource for analyzing AV crash characteristics. Scholars have adopted diverse methodologies, including statistical analysis and validation (Favaro and Varadaraju (2017)), text mining and Bayesian modeling (Boggs and Khattak (2020)), and clustering (Song and Noyce (2021)). However, previous studies often overlooked crash data analysis from the perspective of driving scenarios, especially those leading up to the crash, limiting in-depth insights into the underlying factors and causes of the crash.

Pre-crash scenarios capture critical information about the moments leading up to a crash, defined by features such as the number of vehicles involved, vehicle movements, and crash types. Pre-crash scenarios are instrumental in understanding crash dynamics, key events, and vehicle failure mechanisms (Ren and Feng (2023)). Currently, three main methods are used to identify pre-crash scenarios: pre-crash data analysis (Lee and Yoon (2024)), clustering (Nitsche and Welsh (2017)), and pre-crash scenario typology (Liu and Glaser (2024)). Among these, pre-crash scenario typology provides a systematic classification of scenarios leading up to the crash, based on vehicle kinematics and dynamics characteristics, and key events. Firstly proposed by the National Highway Traffic Safety Administration(NHTSA) in 2007 and revised in 2019, the revised version updated the classification of intersection-related scenarios with greater emphasis on vehicle kinematics and inter-vehicle interactions. Numerous studies have utilized this typology to extract key pre-crash scenarios(Najm and Brewer (2013); Wang and Chen (2022); Lin and Wang (2022); Liu and He (2021)). Among those studies, Liu et al. (Liu and He (2021)) were the first to apply this method to classify AV pre-crash scenarios using 122 AV crash reports from the CA DMV spanning January 2017 and 2020. However, with the development of AV technologies and the increasing scale of AV open road testing, there has been a significant rise in the number of AV crash reports. Furthermore, the typology employed by Liu et al. was based on the old version of the pre-crash scenario typology, underscoring the need for an updated analysis of AV pre-crash scenarios to account for recent advancements and more comprehensive crash data.

[*†]Corresponding Author

Scenario-based crash analysis aggregates crashes into groups based on shared scenario characteristics, enabling the identification of common characteristics of crashes within each scenario while highlighting differences across scenarios, such as influencing factors and causal factors. As autonomous driving technology advances and the scale of road testing expands, AVs encounter increasingly complex scenarios, posing significant safety challenges to AV systems. Consequently, scenario-based AV crash analysis is crucial for enhancing AV safety. Various methods have been used to study crash scenarios, such as Hutton matrix analysis (W. (2015); Wang (2017)), statistical and machine learning models (Haleem and Gan (2015)), correlation analysis (Liu and Glaser (2024); Ashraf and Rahman (2021)), and causal analysis (Yue and Zheng (2020)). However, most related studies lacked a precise definition of crash scenarios, and few focused on specifically defined scenarios. In addition, in-depth causation analysis primarily concentrated on crashes between human-driven vehicles, or between human-driven vehicles with vulnerable road users (VRUs), leaving a gap in research dedicated to in-depth AV crash causation analysis.

To sum up, there are still several areas for improvement in current studies: (1) Using the latest version of pre-crash scenario typology to classify the up-to-date AV crash reports and update the features of AV pre-crash scenarios; (2) Automatic identification method should be proposed to extract the pre-crash scenarios with high efficiency and accuracy; (3) In-depth analysis should be conducted to analyze the underlying causes of AV crashes.

According to the analysis above, this paper makes three contributions to existing research: (1) We collected 618 AV crash reports released by CA DMV up to January 2024 and applied the newly released pre-crash scenario typology to extract scenarios and identify key scenarios. (2) To address the challenges of time-consuming manual recognition and high error rates in scenario extraction, we proposed and validated a set of mapping rules to map crash feature fields to pre-crash scenarios. (3) Based on the AV pre-crash scenarios extracted, this paper inventively combined the method of association analysis and in-depth causation analysis to identify the characteristics of key scenarios, obtain the influencing factors and causal factors, and give optimization suggestions to government and AV manufacturers based on the analysis results.

The remainder of this paper is organized as follows: In section 2, we review previous studies on pre-crash scenario identification and scenario-based characteristics analysis, summarizing their findings and limitations. In section 3, we introduce the data used in this study and conduct a preliminary analysis. In section 4, we outline the methodology of this paper, detailing the logic and indicators of the automatic recognition algorithm for pre-crash scenarios. This section also covers the analysis methods for scenario characteristics, including the use of association analysis to identify crash-influencing factors and the application of an in-depth causal analysis framework for causal investigation. Section 5 presents the results, including scenario extraction outcomes and scenario-based characteristic findings. Finally, section 6 concludes the paper with optimization recommendations and insights for future research.

## 2. Literature review
### 2.1. Pre-crash scenarios identification

Currently, three primary methods are used to construct pre-crash scenarios (Ren and Feng (2023)), including pre-crash data analysis, clustering, and pre-crash scenario typology. Lee et al. (Lee and Yoon (2024)), using California AV collision reports from 2018 to 2022, applied a hierarchical clustering method and combined statistically significant association rules to establish 14 scenarios suitable for AVs. Nitsche et al.(Nitsche and Welsh (2017)) used the k-means clustering method to cluster the historical crash data based on 1056 intersection crash data from On-the-Spot database in the UK, and used the association rules to explore more features of driving scenarios. In this study, 13 categories of T-intersection scenarios and 6 categories of cross-shape intersection scenarios were identified. These studies extracted pre-crash scenarios by analyzing the characteristics of crash data, but the definition of scenarios remained unclear.

Since NHTSA introduced the pre-crash scenario typology, a clear classification method for pre-crash scenarios, many researchers have adopted this more well-defined approach for identifying crash scenarios. Najm et al. (Najm and Brewer (2013)) statistically described 17 pre-crash scenarios using the pre-crash scenario typology for light vehicles in 2013 and for heavy trucks in 2014, respectively. Lin et al. (Lin and Wang (2022)) analyzed 5983 cases from China in-depth traffic accident study database to identify the typical pre-crash scenarios in Chinese road traffic using injuries, road types, accident times, and accident causes as the feature variables. Wang et al. (Wang and Chen (2022)) extracted 536 cases of crashes and near crashes using Shanghai naturalistic driving study and applied pre-crash scenario typology, obtaining 23 different pre-crash scenarios. Liu et al. (Liu and He (2021)) pioneered the use of pre-crash scenario typology in AV crash analysis. They collected 122 AV crashes in California from 2017 to 2020, identifying 15 typical pre-crash scenarios that accounted for a higher percentage, with four pre-crash scenarios highlighted for their high occurrence probability. So far, advancements have been achieved in both AV technology and pre-crash scenario typology. Additionally, more than 600 crash reports have been released by the CA DMV, highlighting the need to update AV pre-crash scenarios and implement more efficient methods for extracting pre-crash scenarios. Moreover, the manual identification and calibration methods in previous studies suffered from high error rates, heavy workloads, and prolonged processing times, highlighting the need for proposing an automated method for pre-crash scenario extraction.

*2.2. Pre-crash scenarios characteristics analysis*

As autonomous driving technology advances, the scenarios that AVs encounter become more complex, and pose significant safety challenges to AV systems, so how to analyze the characteristics of AV crash scenarios has always been the focus. At present, there are several methods for scenario characteristics analysis, such as Hutton matrix analysis, statistical and machine learning models based on process data before, during, and after the accident, etc. Liu et al. (Liu and Glaser (2024)) studied the California AV collision reports from 2014 to 2022 and identified pre-crash scenarios, and then utilized association analysis to recognize the influencing factors in different scenarios. Their study can identify the relationship between crashes and human-vehicle-road-environment features but was limited to intersection-related scenarios. Studies based on association analysis can identify the key influencing factors of specific scenarios, but the correlation is not equivalent to causation.

Therefore, to analyze the explicit causal chain throughout a crash, many scholars have employed in-depth analysis of crash causes, of which driving reliability and error analysis method (DREAM) is one of the most commonly used tools. Yue et al. (Yue and Zheng (2020); Yue and Yuan (2020)) used DREAM to analyze 135 pedestrian crashes in Florida, identified causal factors in five scenarios, and applied association rules to determine the relationship between risk factors and road infrastructures in each scenario. Ye et al. (Ye (2024)) extracted 398 pedestrian crashes from Chinese in-depth accident study and classified the crashes into five scenarios and conducted crash causation analysis for each scenario using DREAM. Wang et al. (Wang and Chen (2022)) proposed a two-stage crash causation analysis method based on pre-crash scenarios and a crash causation derivation framework that systematically categorized and analyzed contributing factors, analyzing the interactions among road users, vehicles, road infrastructures, and road environments. Present research mostly focused on vehicle-pedestrian crash scenarios or traditional vehicle-vehicle scenarios for in-depth causal analysis, while limited studies that utilized causal analysis method for safety assessment of scenarios involving AVs were conducted.

## 3. Data preparation

*3.1. California autonomous vehicle crash reports*

The crash data used in this paper were collected from CA DMV, specifically California AV crash reports spanning from April 2018 to January 2024. The crash report mainly contains five parts: manufacturer's information, information on both parties involved in the crash, personnel and property casualties, and crash description. Consistent with the crash report, in this paper, Vehicle 1 (V1) and Vehicle 2 (V2) represents the AV and the other vehicle involved, respectively.

*3.2. Extraction and coding scheme*

In data preparation, 615 reports were collected initially and were used for further processing and analysis. The main software includes: Python, for PDF information extraction; R, for data processing and analysis, statistical analysis, as well as visualization; and R Studio, for the related processing. The process of crash data extracting and filtering is shown as follows in Figure 1.

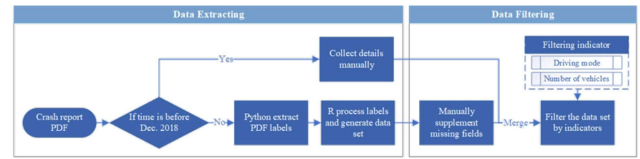

Figure 1: Flowchart of data extracting and filtering

Crash reports prior to 2019 were scanned paper documents and can not be extracted automatically by Python, so manual work was needed to supplement the missing data. The filtering indicators of crash data are: (1) the driving mode of the vehicles (autonomous mode); (2) the number of vehicles involved (2 involved). Finally, 322 crashes were screened and used in the following analysis. The 322 AV crash reports were coded in Figure 2.

About 89% of AV crashes occurred within intersection or intersection-related areas (Favaro and Varadaraju (2017)), so these two types of areas are important for intersection crash safety analysis (Wang and Chen (2022)). Therefore, in this paper, we adopt the same classification with previous study(Liu and Glaser (2024)), classifying the intersection area into intersection impact area and intersection center.

*3.3. Basic analysis of crash characteristics*

*3.3.1. Other parties involved*

Beside vehicle-vehicle crash, the 322 crashes involved other traffic participants including non-motorized vehicles, pedestrians, animals, and objects. The annual distribution of other parties involved in crashes was analyzed in Figure 3.

In Figure 3, each ring represents one year, with different colors representing different types of traffic participants involved.

Figure 2: Coding scheme of AV crash report

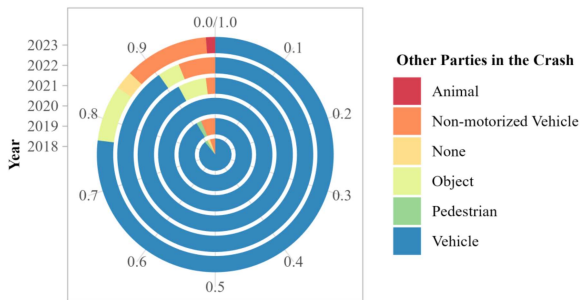

Figure 3: Annual distribution of other parties involved

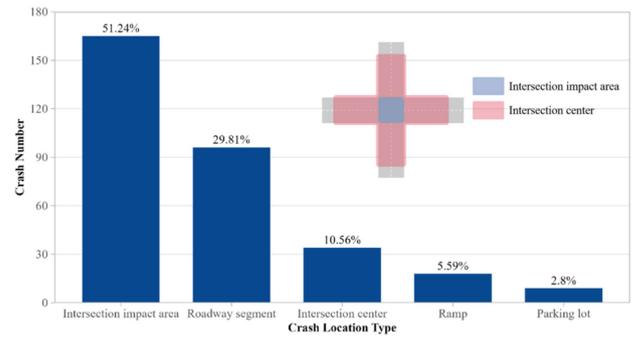

Figure 4: Distribution of crash location types

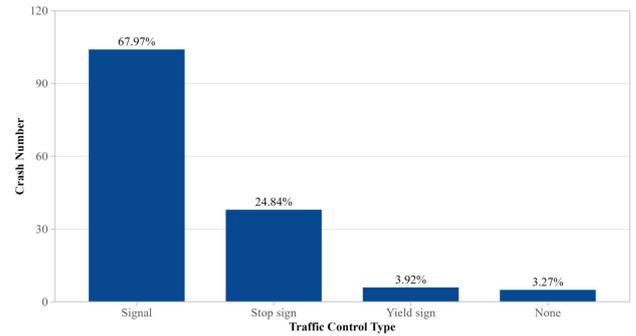

Figure 5: Distribution of traffic control types in the intersection area

Moving outward from the central circle, the year incrementally increases. The majority of AV crashes involve motorized vehicles; however, it is evident that the proportion of crashes involving non-motorized vehicles and objects increases rapidly. In 2023, the number of crashes related to non-motorized vehicles and objects exceeds 20%. It is presumed that as the scale and scope of AVs' open road testing increase, they encounter more complex and diverse traffic conditions, thereby increasing the probability of crashes with other traffic parties.

### 3.3.2. Traffic environment

The location types of crashes include roadway segment, intersection impact area, intersection center, parking lot, and ramp, as shown in Figure 4. As illustrated in Subsection 3.2, the intersection center refers to the area where traffic flow from different directions weaves within the intersection area, and the intersection impact area refers to the region of approaches to the intersection center. For the 199 crashes that occurred in the intersection area, the types of intersection control were analyzed in Figure 5.

The crashes mainly happened in the intersection impact area, accounting for more than 50%, followed by the roadway segment, accounting for nearly 30%, and there was also 11% of crashes occurred in the intersection center. It can be found that the intersection area, including the intersection impact area and intersection center, is the main location of AV crashes. Further analysis of the traffic control types in the intersection area reveals that most of the intersections are signal controlled (67.97%), followed by stop sign control (24.84%). Therefore, in subsequent analysis, it is necessary to focus on signal controlled as well as stop sign controlled intersections.

### 3.3.3. Damage severity

Statistical analysis of AV crash damage severity found that most of the crashes were slightly damaged (75.5%), followed by moderate damaged (14%), and the proportion of severely damaged crashes reached 2.8%. The heat map of AV damage location distribution is plotted in Figure 6, where the number in each region represents the frequency of damage in that region.

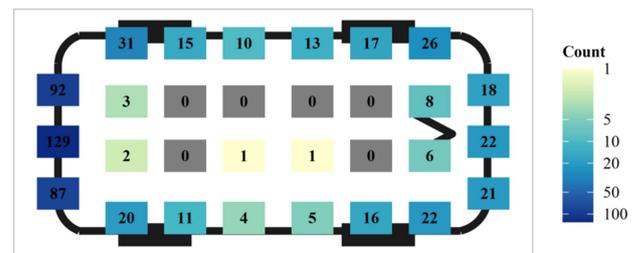

Figure 6: AV damage location distribution heat map

The rear bumper of AV and other regions of the rear had the highest damage frequency, indicating that the probability of AV crashes being rear-ended by other vehicles was higher. The higher frequency of damage to the left/right rear suggested that other vehicles may rear-end the AVs in front of them while changing lanes or overtaking, highlighting the need to lay more emphasis on this issue.

## 4. Methodology

### 4.1. Pre-crash scenario identification

Pre-crash scenario typology is a classification method describing vehicle kinematics and dynamics features, as well as key events that occur prior to the crash. The typology has been optimized in several versions since first introduced in 2003. In 2019, a newly revised version was published (36 pre-crash scenarios) (Elizabeth and Azeredo (2019)), which focuses more on vehicle dynamics and kinematics characteristics and primarily optimizes the classification of intersection-related scenarios in Figure 7, providing a better understand of the vehicle interaction in intersection areas. Therefore, in this paper we adopt this latest version of pre-crash scenario typology.

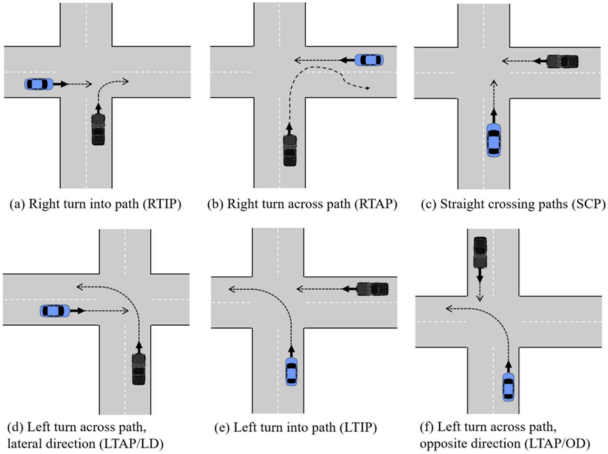

Figure 7: Schematic of intersection scenarios

In this paper, automatic mapping of pre-crash scenarios from feature fields was achieved by adding relevant feature fields and setting scenario mapping rules based on vehicle kinematics and scenario spatial features. We used supplemented feature fields identified through the crash description in the crash reports (i.e. vehicle dynamics, vehicle kinematics, vehicle relative position, vehicle speed variation, etc.), together with the fields automatically extracted from the crash reports as the mapping fields. Then mapping fields are sorted and encoded, as shown in Figure 8.

Considering the diversity of vehicle motions and relative positions, there may be multiple sub-scenarios for each scenario, so the initial mapping rules according to a standard scenario definition may overlook some sub-scenarios. Therefore, we used R to execute the mapping rules, combined with manual check to find the missed sub-scenarios. Then we revised or added mapping rules, and iterated this process to maximize the identification accuracy, as depicted in Figure 9.

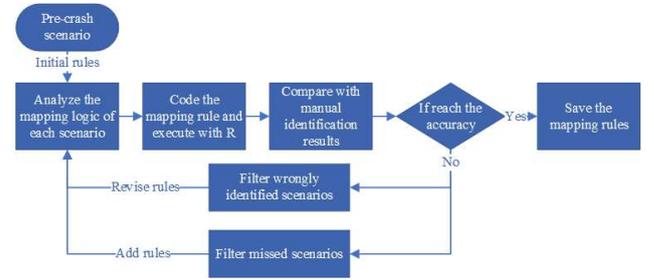

Figure 8: Encoding values of mapping fields

Figure 9: Construction of scenario mapping rules

The automatic identification of pre-crash scenarios is assessed using indicators such as false alarm rate (FAR), missing alarm rate (MAR), and accuracy. These indicators are calculated using Equation 1, Equation2, and Equation 3.

$$FAR = \frac{FICN}{TCN} \quad (1)$$

$$MAR = \frac{\sum_{i=1,2,...,36} MICN_i}{TCN} \quad (2)$$

$$Accuarcy = \frac{CICN}{TCN} \quad (3)$$

Where, $FICN$ is the falsely identified crash number; $TCN$ means the total crash number; $MICN$ represents the missed identified crash number; $i$ is the scenario index; and $CICN$ is the correctly identified crash number.

### 4.2. Characteristic analysis of pre-crash scenarios

Association rules and Driving Reliability and Error Analysis Method (DREAM) are used to analyze the characteristics of the key scenarios.

#### 4.2.1. Association rules

Currently, the most widely applied association analysis method in related studies is the Apriori algorithm (Datu (2023)), therefore, this study adopted the Apriori algorithm to analyze the potential connections between scenarios and environmental characteristics. Three key parameters are used for identifying and filtering strong association rules, including support, confidence, and lift.

Support is used to measure the frequency of item set in the total entries. In this study, we paid more attention to rules with greater support. We calculate the support of rule $X \Rightarrow Y$ using Equation 4.

$$Support(X \Rightarrow Y) = P(X \cap Y) = \frac{N(X \cap Y)}{N} \quad (4)$$

Where, $N(X \cap Y)$ is the number of entries that contains both item set $X$ and $Y$; and $N$ is the total number of entries.

Confidence is used to measure the proportion of frequency that item set $X$ and $Y$ co-occur relative to the frequency of item set $X$. A higher confidence indicates that under the occurrence of the premise $X$, the probability of occurrence of $Y$ is higher. The confidence of rule $X \Rightarrow Y$ can be obtained in Equation 5.

$$\begin{aligned} Confidence(X \Rightarrow Y) &= \frac{Support(X \Rightarrow Y)}{Support(X)} \\ &= P(Y|X) \\ &= \frac{N(X \cap Y)}{N(X)} \end{aligned} \quad (5)$$

Where $N(X)$ is the number of times the item set appears in total entries.

Lift is used to measure the independence between item set $X$ and $Y$. The equation for calculating lift of the rule is shown in Equation 6.

$$\begin{aligned} Lift(X \Rightarrow Y) &= \frac{Support(X \Rightarrow Y)}{Support(X) \cdot Support(Y)} \\ &= \frac{P(X \cap Y)}{P(X) \cdot P(Y)} \\ &= \frac{N(X \cap Y) \cdot N}{N(X) \cdot N(Y)} \end{aligned} \quad (6)$$

When lift is equal to 1, it means that the two item sets are not related; when lift is smaller than 1, it means that item sets are mutually exclusive; when lift larger than 1, it is considered that there is a positive correlation between the occurrence probability of two item sets.

The minimum length of the rules was set as three and the maximum length as six, and 342153 rules were initially generated. Through three key parameters and the previous analysis results on the AV crashes characteristics, the significant rules with strong association were screened for each type of scenarios.

#### 4.2.2. Driving reliability and error analysis method

The DREAM is a structured analytical method to derive the causal chain of a crash and identify the key causal factors of the crash. DREAM deduces the casual chain of a crash by a set of formally defined classification scheme (causation factors, genotypes and phenotypes) and links (causation relationships among the causation factors). The causal analysis chain of the DREAM method is shown in Figure 10. The classification scheme of DREAM comprises a number of observable effects in the form of human actions and system events called phenotypes.

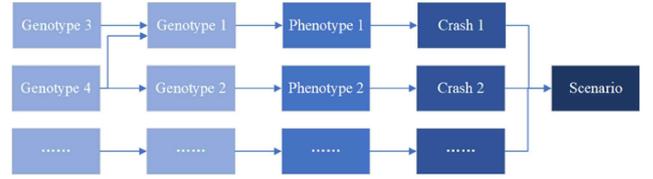

Figure 10: Causal chain of DREAM

## 5. Results and discussion

The comprehensive accuracy of automatic scenario identification is obtained in Table 1.

Table 1: Automatic scenario identification accuracy

| Indicator | Value |
| --- | --- |
| Number of scenarios identified correctly | 21 |
| False alarm rate (%) | 1.9 |
| Missing alarm rate (%) | 0.6 |
| Accuracy (%) | 98.1 |

The mapping rules can identify most scenarios with complete accuracy (87.5%), of which the scenarios that require manual check mainly include vulnerable road user (VRU) related scenarios. In NHTSA pre-crash scenario typology, there are fewer pre-crash scenarios of VRUs, and pedestrians as well as non-motorized vehicles are highly mobile. However, there is no specific classification of VRUs' motions in the crash reports, and their behaviors are mentioned only in the detailed description, which leads to inaccurate identification of VRU related scenarios.

The false alarm rate of automatic identification is 1.9% (6 cases out of 322 cases), which is mainly due to the inaccurate judgment of complex scenarios. For example, if both vehicles are changing lanes before the crash, the exact scenario needs to be further confirmed with detailed descriptions, yet using mapping rules may lead to inaccurate results. The missing rate of 0.6% is primarily attributed to the incomplete recognition of the other scenario (No. 36 scenario). The overall scenario identification accuracy is 98.1%, which can accurately identify the scenarios of the vast majority of crashes. Based on the aforementioned indicators, the proposed automatic mapping rules can function effectively.

### 5.1. Pre-crash scenario mapping

For the 322 crashes studied, 24 pre-crash scenarios are extracted. The scenarios are counted and arranged in decreasing order in Table 2. The serial numbers of the scenarios in the table are consistent with the serial numbers in the NHTSA pre-crash scenario topology.

Table 2: Pre-crash scenarios of AVs

| No. | Pre-crash scenario | Count | Percentage |
|---|---|---|---|
| 24 | Rear-end/Lead Vehicle Stopped (LVS) | 105 | 32.61 |
| 20 | Rear-end/Following Vehicle Making a Maneuver and Approaching Lead Vehicle (FVM) | 38 | 11.80 |
| 23 | Rear-end/Lead Vehicle Decelerating (LVD) | 35 | 10.87 |
| 13 | Backing into Vehicle | 19 | 5.90 |
| 12 | Pedalcyclist/No Maneuver | 17 | 5.28 |
| 17 | Drifting/Same Direction | 13 | 4.04 |
| 19 | Opposite Direction/No Maneuver | 13 | 4.04 |
| 15 | Parking/Same Direction | 11 | 3.42 |
| 22 | Rear-end/Lead Vehicle Moving at Lower Constant Speed (LVM) | 7 | 2.17 |
| 28 | Left Turn Across Path, Lateral Direction (LTAP/LD) | 7 | 2.17 |
| 35 | Object/No Maneuver | 7 | 2.17 |
| 36 | Other | 7 | 2.17 |
| 27 | Straight Crossing Paths (SCP) | 7 | 2.17 |
| 16 | Changing Lanes/Same Direction | 7 | 2.17 |
| 14 | Turning/Same Direction | 7 | 2.17 |
| 21 | Rear-end/Lead Vehicle Accelerating (LVA) | 6 | 1.86 |
| 34 | Object/Maneuver | 5 | 1.55 |
| 12 | Pedalcyclist/Maneuver | 3 | 0.93 |
| 30 | Left Turn Across Path/Opposite Direction (LTAP/OD) | 2 | 0.62 |
| 1 | Vehicle Failure | 2 | 0.62 |
| 8 | Animal/No Maneuver | 1 | 0.31 |
| 29 | Left Turn Into Path (LTIP) | 1 | 0.31 |
| 33 | Non-collision/No Impact | 1 | 0.31 |
| 10 | Pedestrian/No Maneuver | 1 | 0.31 |

The majority of identified scenarios are scenarios between vehicles, with the most frequent scenarios in the order of rear-end/lead vehicle stopped (LVS) (32.61%), rear-end/following vehicle making a maneuver (FVM) (11.8%) and rear-end/lead vehicle decelerating (LVD) (10.87%). The results are similar to related studies (Liu and He (2021)) that the highest percentage of scenarios obtained is LVS scenario. Due to the high percentage of rear-end scenarios, the above three types of rear-end scenarios will be taken as the key scenarios and analyzed in-depth in the following characteristics analysis section.

As can be seen in Figure 7, the intersection scenarios totaled 17 crashes, accounting for 5.27%. However, the analysis of crash severity reveals that AVs in intersection scenarios are more seriously damaged, with 40% of them being moderately and above damaged, of which 20% are severely damaged. Given that most of the existing intersection crashes are severe crashes and that one of the main challenges is how to operate safely and efficiently under complex traffic conditions such as intersections, so intersection scenarios are also key scenarios that need to lay more focus on.

In summary, considering the high percentage of the three types of rear-end scenarios and the severity of the crashes in intersection scenarios, those two are therefore taken as the key scenarios in the subsequent section of this paper for further analysis of scenario characteristics.

### 5.2. Characteristic analysis of the key scenarios

Three specific types of rear-end scenarios and intersection scenarios will be discussed in this section. It should be noted that whilst some of the crashes in rear-end scenarios occur in intersection areas, in this paper, the rear-end scenarios are defined as crashes between a following vehicle and its preceding vehicle, i.e., crashes between vehicles in the same travel direction, while the intersection scenarios are crashes between vehicles of different travel directions in intersection areas as can be seen in Figure 11.

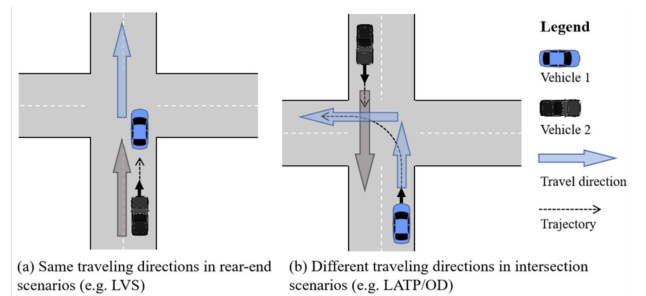

(a) Same traveling directions in rear-end scenarios (e.g. LVS)  (b) Different traveling directions in intersection scenarios (e.g. LATP/OD)

Figure 11: Vehicle travel direction in different scenarios

Given that the crash reports for rear-end scenarios contain more environmental information about the crash location and fewer details about the crash process, the crash reports for intersection scenarios contain detailed information about the crash process, allowing for in-depth analysis of the crash causation.

Therefore, this paper employs the method of combining association analysis and causation analysis. Association analysis is conducted for the rear-end scenarios with relatively less information, to explore the potential association between different rear-end scenarios and the traffic environment, and to identify the key influencing factors. DREAM causation analysis is applied for the intersection scenarios with more detailed information, to analyze the phenotypes and genotypes in-depth, and to sort out the causation chain of the scenarios, thus, identifying the significant causal factors.

*5.2.1. Characteristics Analysis of Rear-end Scenarios*

The key scenarios identified include three types of rear-end scenarios (i.e., LVS, FVM, and LVD). Typical schematics of the three types of focused rear-end scenarios are shown in Figure 12.

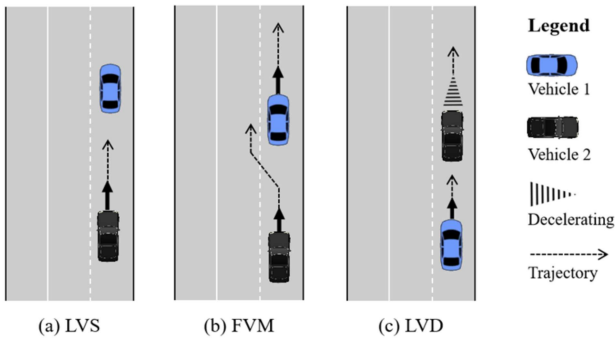

Figure 12: Schematics of three types of key rear-end scenarios

*Lead vehicle stopped scenario.* After filtering the duplicate rules using lift and support, significant association rules for LVS scenario are obtained and arranged in descending order according to support, as can be seen in Table 3.

Table 3: Association rules for LVS scenario

| No. | Rules | Support | Confidence | Lift |
| --- | --- | --- | --- | --- |
| 1 | {Location_Type=Intersection impact area + Traffic.Control.Type=signal + V1.intention=signal stopped} | 0.168 | 0.969 | 1.73 |
| 2 | {Location_Type=Intersection impact area + Traffic.Control.Type=signal + V1.intention=signal stopped+Lane.markings=lane markings} | 0.158 | 0.967 | 1.73 |
| 3 | {V1.intention=right turn&yield + Cycle.lane=no-no cycle lane} | 0.071 | 0.929 | 1.66 |
| 4 | {V1.intention=right turn&yield + Roadside.parking=no + Lane.markings=lane markings} | 0.071 | 0.929 | 1.66 |
| 5 | {V1.intention=right turn&yield + Type.of.intersection=cross-shaped} | 0.065 | 1 | 1.79 |
| 6 | {Location_Type=Intersection impact area + Traffic.Control.Type=stop sign + V1.intention=stopped stop sign + Type.of.intersection=cross-shaped} | 0.043 | 1 | 1.79 |
| 7 | {If_peak_time=non-peak + Location_Type=Ramp + V1.intention=merging&yield + Roadside.parking=no} | 0.038 | 1 | 1.79 |
| 8 | {Location_Type=Intersection impact area + V1.intention=left turn&yield + V1.yield.for=oncoming traffic + Cycle.lane=no-no cycle lane} | 0.022 | 1 | 1.79 |
| 9 | {Location_Type=Intersection impact area + V1.intention=left turn&yield + V1.yield.for=oncoming traffic + Roadside.parking=yes + Cycle.lane=no-no cycle lane} | 0.022 | 1 | 1.79 |
| 10 | {Location_Type=Intersection impact area + Traffic.Control.Type=signal + V1.intention=left turn&yield + V1.yield.for=oncoming traffic + Cycle.lane=no-no cycle lane} | 0.022 | 1 | 1.79 |

Rules 1 and 2 show that the most frequent environmental features in LVS scenario are: intersection impact area, signal control, lane markings, and common AV intention is stopping at signal; rules 3-5 show that when the AV intends to turn right, the environmental features that occur together include no roadside parked vehicles, no non-motorized traffic lanes, and an intersection; rule 6 shows that it is prone to occur LVS scenario when AV stops at a stop sign at intersection impact area; rule 7 indicates that when AV waits to merge and yields to main-road traffic on ramps in off-peak times, LVS scenario is prone to appear; and rules 8-10 indicate that AV is prone to be involved in LVS scenario when turning left and yielding to the oncoming traffic at signal controlled intersection impact area, under the condition of parked vehicles on the roadside. Visualization of the top 10 rules for LVS scenario is plotted in Figure 13.

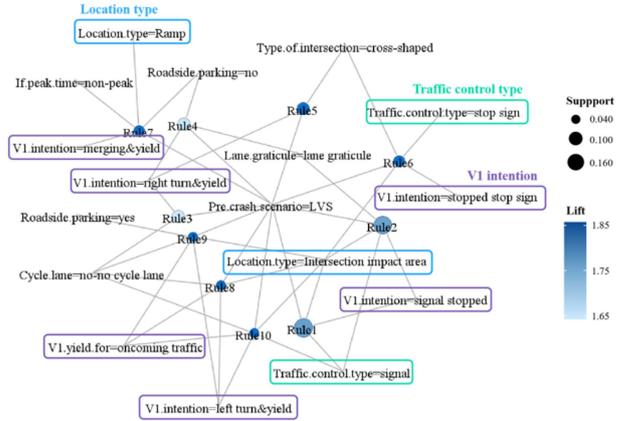

Figure 13: Visualization of the top 10 rules for LVS scenario

It can be seen from Figure 13 that the significant influencing factors of LVS scenario include the location type (i.e., ramp, and intersection impact area), traffic control type (i.e., stop sign control, and signal control), and the AV intention (i.e., stop at a signal, stop at a stop sign, stop to yield to oncoming traffic when turning left/right, etc.).

*Following vehicle making maneuver scenario.* After filtering, we can obtain the significant association rules for FVM scenario and rearrange them according to support in Table 4.

Table 4: Significant association rules for FVM scenario

| No. | Rules | Support | Confidence | Lift |
| --- | --- | --- | --- | --- |
| 1 | {Location_Type=Roadway segment + V1.intention=proceed straight + Roadside.parking=yes + Lane.markings=lane markings} | 0.043 | 1 | 4.49 |
| 2 | {Location_Type=Roadway segment + Roadside.parking=yes + Number.of.lanes.one.direction=3 + Lane.markings=lane markings} | 0.043 | 1 | 4.49 |
| 3 | {Roadside.parking=yes+Road.types=one-way + Number.of.lanes.one.direction=2 + Lane.markings=lane markings} | 0.033 | 0.857 | 3.85 |
| 4 | {If_peak_time=non-peak + Location_Type=Roadway segment + V1.intention=stopped in road traffic + Roadside.parking=yes + Lane.markings=lane markings} | 0.027 | 1 | 4.49 |
| 5 | {Lighting=dark-street lights + Location_Type=Roadway segment + Roadside.parking=yes + Number.of.lanes.one.direction=3 + Lane.markings=lane markings} | 0.027 | 1 | 4.49 |

Rule 1 indicates that the high frequency occurrence characteristics of FVM scenario are: on roadway segment with lane markings and roadside parking, and the AV state is straight ahead. When the AV intention is not considered, from rules 2, 3 and 5, it is found that the most frequent environmental factors for FVM scenario include: nighttime, roadway segment, road side parking, and number of one-way lanes is 2 or 3 with lane markings. According to rule 4, FVM scenario also occurs when AV follows the roadway traffic and stops on a roadway segment with lane markings and roadside parking. Visualization of the significant rules for FVM scenario is plotted in Figure 14.

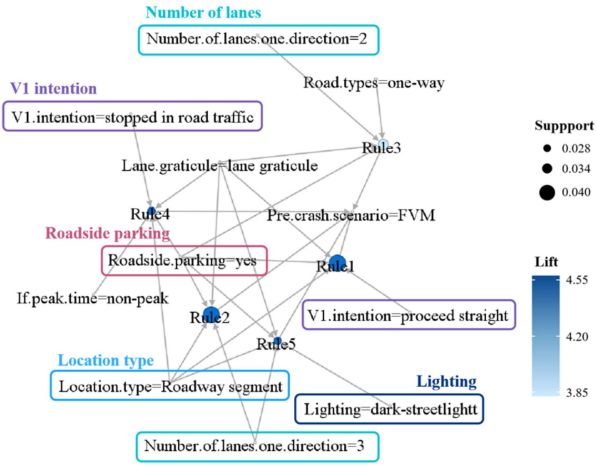

Figure 14: Visualization of significant rules for FVM scenario

It can be found that the influencing factors related to FVM scenarios mainly include the location type (i.e., road section), light (i.e., at night), number of lanes (i.e., lanes number is 2 or 3), roadside parking (i.e., the existence of roadside stopped vehicles), and AV intention (i.e., slow in the traffic flow, and go straight).

*Lead vehicle decelerating scenario.* The significant association rules for LVD scenario are obtained and sorted in Table 5.

Table 5: Significant association rules for LVD scenario

| No. | Rules | Support | Confidence | Lift |
|---|---|---|---|---|
| 1 | {Lighting=c + V1.intention=proceed straight&yield + Lane.markings=lane markings} | 0.033 | 0.857 | 4.51 |
| 2 | {Lighting=c + V1.intention=proceed straight&yield + Roadside.parking=yes + Lane.markings=lane markings} | 0.027 | 1 | 5.26 |
| 3 | {V1.intention=proceed straight&yield + Cycle.lane=yes-no separation + Lane.markings=lane markings} | 0.027 | 0.833 | 4.38 |
| 4 | {V1.intention=proceed straight&yield + Road.types=two-way with marked median + Lane.markings=lane markings} | 0.027 | 0.833 | 4.38 |
| 5 | {Traffic.Control.Type=signal+V1.intention=proceed straight&yield + Cycle.lane=yes-no separation + Lane.markings=lane markings} | 0.027 | 0.833 | 4.38 |
| 6 | {Location_Type=Intersection impact area + Type.of.intersection=T-shaped + Road.types=two-way with marked median + Lane.markings=lane markings} | 0.022 | 0.8 | 4.21 |

From rules 1 and 2, it can be found that LVD scenarios often occur at night, with the road having lane markings and roadside parking, and AV is rear-ended while going straight and decelerating. It can be seen from rules 3-5 that the high-frequency environmental features of LVD scenario are: the road has lane markings and non-motorized lanes without marking separation, and the type of the road is a bi-directional carriageway with a center line. In addition, from rule 6, it is found that LVD scenarios occur relatively more frequently in the T-shaped intersection impact area. Visualization of the top 6 rules for FVM scenario is plotted in Figure 15.

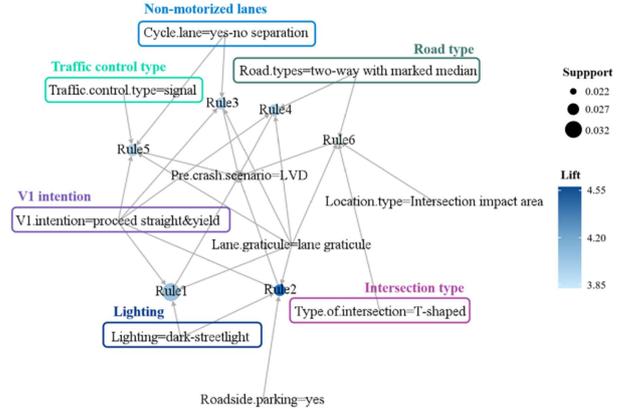

Figure 15: Visualization of significant rules for LVD scenario

It can be found that the key influencing factors related to LVD scenario include the traffic control type (i.e., signal control), light (i.e., at night), road type (i.e., two-way lanes separated by a median line), intersection type (i.e., T-shaped intersection), non-motorized lanes (i.e., non-motorized lanes without segregation), and the AV intention (i.e., slowing down to yield to the vehicle ahead when travelling straight).

In conclusion, the environmental influencing factors varied in different types of rear-end scenarios: the most significant factors in LVS scenario are mainly location type, including signal/stop sign-controlled intersection impact area, and ramps; the most significant factors in FVM scenario included road type and light, with the road type being mainly roadway segments with lane markings and roadside parking, and the scenario appears more frequently at night with street light; the most significant factors in LVD scenario are lighting, roadside parking and non-motorized lanes.

*5.2.2. Characteristics Analysis of Intersection Scenarios*

In this paper, we analyze the behaviors of conventional vehicles during 15 intersection crashes (excluding two crashes with ambiguous locations), based on DREAM 3.0 (Warner and Johansson (2008)), calculate causal factors, and plot the causal chain of intersection scenarios Figure 16. The crash number of the corresponding factor is marked in parentheses. The green boxes are phenotypes, and the blue boxes are genotypes.

*Phenotype analysis.* It can be seen that the phenotype includes four types: timing, speed, distance, and direction, among which the timing phenotype accounts for the largest proportion, including two specific phenotypes (i.e., too early action and no action).

*Genotype analysis.* According to the DREAM framework, the genotypes are categorized by driver, vehicle, traffic environment, and organization. Yet the vehicle genotypes are not included in this study since minimal details are contained about the conventional vehicle in the crash reports (e.g. vehicle illumination problems, equipment failure, etc.).

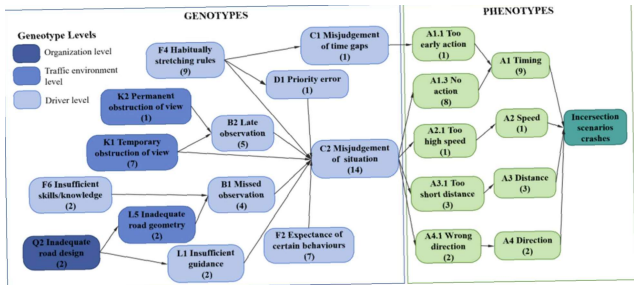

Figure 16: Causal chain of intersection scenarios

Genotypes that directly contribute to the phenotype are driver's interpretation, including misjudgment of time gaps and misjudgment of situation, and the latter one is the main causal factor. In the permanent personal level, three genotypes are identified: expectancy of certain behaviors, habitually stretching rules, and insufficient skills/knowledge. When analyzing the genotypes in-depth, we find that in the intersection scenarios, 67% (10 cases) of the crashes occurred at night, which is significantly higher than that in the overall crashes and rear-end scenarios, 32.3% and 26.4% respectively. Drivers may therefore be motivated by the expectation that there are fewer vehicles at night, leading to misjudgment of the situation. In addition, the habitual violation factor occurs more frequently (60%), including five running stop sign cases and four running red light cases.

Under observation level includes two genotypes: missed observation and late observation. The potential causes of these genotypes include two parts: the environmental effect and the driver's personal skill deficiency. At the environmental level, genotypes identified are permanent obstruction of view, such as vegetation in the central green belt, and temporary obstruction to view, e.g. parked vehicles on both sides of the road, which can block the driver's view and affect his/her judgment. Based on the number in parentheses, it can be found that 50% of the crashes had temporary obstruction of view as one genotype of the causal chain, and all of them are obstructed by roadside parked vehicles. In addition, insufficient skills/knowledge can also affect driver observation, thus impact their judgment of the situation, such as driving the wrong way due to unfamiliarity with the road.

The state of the road affects drivers in terms of misjudging the situation due to insufficient guidance, or not perceiving due to insufficient road geometry. In two crashes, the two-way road is wide with no central lane marking, in an X-type intersection, so it is not easy for human drivers to judge the distance when performing turning and other operations, resulting in collisions within the intersection center.

Based on the analysis above, it can be found that among the genotypes leading to crashes in intersection scenarios, habitual violations and expectations of certain behaviors are the key causal factors leading to intersection scenario crashes, followed by temporary obstruction of view. Habitual violations are also associated with drivers taking chances due to the low traffic volume at night. At the environmental level, temporary obstruction of view due to roadside parking is also a key causal factor. Based on the safety assessment results of the key scenarios, this paper proposes optimization recommendations from both the perspectives of government authorities and AV manufacturers.

## 6. Conclusion

### 6.1. Recommendations

#### 6.1.1. Government

The results of LVS scenario assessment indicate that high-grade ramps are frequently the site of crashes. With the street view in Google Maps on the crash site, we find most crashes occurred in the circular region of the ramp with tall greenery blocking the drivers' sight. Therefore, appropriate pruning of the roadside greenery is needed, so that drivers have enough vision and reaction time to the road conditions ahead. Based on the intersection scenario analysis results, the government needs to reasonably divide the roadside parking spaces in the intersection areas to prevent parked vehicles from obstructing the driver's view. Furthermore, there is a need to increase the supervision of drivers' illegal behavior at night, as well as the need to educate drivers to avoid taking chances.

#### 6.1.2. Manufacturer

According to the association results of rear-end scenarios, when conducting open road testing, AV manufacturers need to lay more attention on scenarios such as ramps between high-level roads, stop sign/signal-controlled intersections, road sections with roadside parking, and T-shaped intersections; in terms of lighting conditions, it is necessary to pay attention to scenarios at night with street light. The analysis results of intersection scenarios reveal that AVs may have insufficient perception of the potential collision risks at the rear of the vehicle, or there may be no effective avoidance measures for the perceived collision risks. So consideration can be given to adding other risk-avoiding emergency maneuvers, such as moving forwards, changing lanes, etc., to reduce the collision risks at the rear of the vehicle, when there is spare space. Also, currently AVs all choose to decelerate to a complete stop to avoid the collision with the preceding vehicle (most crashes in LVS scenario). This may be inconsistent with the human drivers' anticipation of AVs movements, coupled with the fact that AVs' reaction decision-making time is smaller than that of human drivers, and the latter may not be able to synchronize their reaction when AVs are performing emergency braking, which causes the conventional vehicle rear-ending the front AV. Therefore, AV manufacturers may consider appropriately adjusting the conservativeness of AVs to adapt to the current mixed traffic flow.

### 6.2. Conclusion and future work

At a time when AV safety is being challenged, one of the most efficient ways to develop AV optimization and recommendations is to identify the influencing factors as well as the causal factors of AV crashes based on pre-crash scenarios. This paper studies the latest 322 California autonomous vehicle collision reports and obtains the crash distribution and damage severity characteristics through preliminary statistical visualization. Based on the latest NHTSA pre-crash scenario typology, this paper designs a set of automatic mapping rules using vehicle dynamics and kinematics, as well as spatial feature fields, which realizes automatic identification of 24 pre-crash scenarios with a 98.1% accuracy. By analyzing the proportion and severity of the scenarios, two types of key scenarios are obtained, namely, high-frequency rear-end scenarios and intersection scenarios with severe crashes. The characteristics analysis of these two key crash scenarios takes into account the data feature, and adopts association analysis and causation analysis, respectively, to obtain the key influencing factors of different rear-end scenarios, as well as the causation chain and key causal factors of the intersection scenarios. Finally, improvement suggestions based on the results of the safety assessment are proposed, both from the perspective of the government and AV manufacturers. This paper highlights the importance of AV pre-crash scenarios in understanding the causes and influencing factors of crashes. The research can guide the construction of AV test scenarios and the optimization of AV control algorithms to improve the safety of AV in real-world traffic environments and can also help the public understand the current state of AV safety.

In the future, this paper will be further deepened in the following aspects: (1) In this paper, the analysis of the key pre-crash scenarios is constrained to crashes between AVs and conventional vehicles, whereas the number of crashes among AVs and non-motorized vehicles has been increasing rapidly in recent years. Since pedestrians and non-motorized vehicles are the vulnerable parties of road users, increased attention needs to be given to crashes between AVs and VRUs. (2) In the analysis of scenario characteristics, limited to the black-box feature of the automated driving algorithms, it is not possible to use DREAM causal analysis to structure the perception-decision-execution process of AVs during crashes. (3) In addition, no corresponding simulation has been built to validate the optimization scheme proposed in this paper that the conservativeness of AVs should be adjusted to adapt to the current mixed traffic flow, so further validation of the effectiveness of the optimization scheme is needed in our subsequent studies.

## Acknowledgements

The authors have stated that they have no potential conflicts of interest regarding the research, authorship, and publication of this article. Furthermore, they have confirmed that they did not receive any financial support for the research, authorship, and publication of this article.